\documentclass[11pt]{article}
\usepackage[margin=1.1in]{geometry}
\usepackage{amsmath,amssymb}
\usepackage{graphicx}
\usepackage[colorlinks=true,linkcolor=blue,citecolor=blue,urlcolor=blue]{hyperref}
\usepackage{microtype}

\newcommand{\tstar}{t^{*}}
\newcommand{\Droll}{\mathrm{Roll}_{\theta}}

\title{Twin Rollouts: Noise-Coupled Counterfactual Branching\\in Interactive Video World Models}
\author{Yu Ma \and Hongli Shi \and Xinran Xu}
\date{}

\begin{document}
\maketitle

\begin{abstract}
Interactive video world models generate rollouts autoregressively under an
action stream, yet they are trained and evaluated almost exclusively on factual
prediction. We study counterfactual generation \emph{inside} the rollout: given
a trajectory the model has itself generated, what would have happened had the
actions differed from step $\tstar$ onward? We formalize \textbf{noise-coupled
twin rollouts} --- a factual and a counterfactual branch sharing the generated
prefix and the future exogenous noise sequence, diverging only in the action
stream at an intervention point. Because the factual branch is self-generated,
its exogenous noise is known exactly: the abduction step of Pearl's
counterfactual procedure is \textbf{exact by construction}, sidestepping the
approximate-inversion problem faced by editing-based pipelines. Noise coupling
further turns the minimal-change principle into a \textbf{per-sample verifiable
property}: we define a spatiotemporal locality metric that penalizes divergence
outside the causal descendants of the intervention, computable against
simulator ground truth without a learned judge. Forking the simulator state at
$\tstar$ yields ground-truth counterfactual re-renders, which we use as
\textbf{verifiable rewards} for post-training. This note establishes the formal
framework, metric definitions, and positioning; experiments are forthcoming.
\end{abstract}

\section{Introduction}

Interactive video world models roll forward autoregressively under a stream of
actions. Their training objectives, and nearly all of their evaluations, concern
the \emph{factual} rollout: how faithfully the model continues the trajectory it
is on. This note concerns a different question, asked from inside the rollout:
had the actions differed from some step $\tstar$ onward, what would this very
trajectory have become? We make three contributions.

\paragraph{(C1) In-rollout counterfactual branching.}
To our knowledge, we give the first formalization of \emph{generation-time
counterfactual branching} for autoregressive interactive video world models:
counterfactual branches share the self-generated factual prefix and the future
exogenous noise sequence, diverging only in actions from an intervention point
$\tstar$. The construction is \textbf{abduction-free}: unlike editing-based
counterfactuals that approximately invert an observed video, the exogenous noise
of a self-generated trajectory is stored, not inferred.

\paragraph{(C2) Per-sample verifiable locality.}
Under noise coupling, any factual--counterfactual divergence outside the causal
descendants of the intervention is attributable to a locality violation rather
than to sampling stochasticity. We define a spatiotemporal locality metric using
simulator-provided descendant masks: divergence must (i) begin only at $\tstar$
and (ii) remain confined to the descendant region as it propagates. The metric
requires no learned judge and is directly reusable as a training signal.

\paragraph{(C3) Counterfactual-pair verifiable rewards.}
Forking the simulator state at $\tstar$ yields ground-truth counterfactual
re-renders. We combine counterfactual outcome fidelity (against the forked
ground truth) with locality (C2) into verifiable rewards for RL post-training
--- to our knowledge the first post-training objective that directly optimizes
the \emph{counterfactual correctness} of a video world model, as opposed to
reconstruction, perceptual, geometric, or action-following rewards.

\paragraph{Separability.}
These contributions are separable: the locality metric (C2) and the
counterfactual-reward objective (C3) do not presuppose the branching formalism
(C1) and stand on their own.

\section{Formal framework}

\begin{description}
\item[Definition 1 (Interactive rollout with explicit noise).]
A world model generates $x_{k+1} = G_{\theta}(x_{\le k}, a_k, \varepsilon_k)$,
where $\varepsilon_k$ collects all exogenous randomness consumed at step $k$ ---
for diffusion/flow backbones, the initial latent noise and any stochastic-solver
noise of chunk $k$; for autoregressive token backbones, the per-position
sampling variates (inverse-CDF uniforms or, equivalently, Gumbel keys) of the
tokens of step $k$. A rollout
$\tau = (x_0, a_{0:T-1}, \varepsilon_{0:T-1}, x_{1:T})$ is fully determined by
$(x_0, a, \varepsilon)$ given $\theta$.

\item[Definition 2 (In-rollout intervention; noise-coupled twin).]
Given a factual rollout $\tau^{F}$ and $\tstar \in \{1,\dots,T-1\}$, an
intervention is $I = \mathrm{do}(a_{\tstar:T-1} := a'_{\tstar:T-1})$ (action
edit) or $\mathrm{do}(x_{\tstar} := x')$ (state edit). The noise-coupled
counterfactual branch is
$\tau^{CF} = \Droll(x^{F}_{\le\tstar}, a', \varepsilon^{F}_{\tstar:T-1})$:
same prefix, same future noise, different actions. Exogenous noise is indexed by
absolute timestep: the counterfactual branch consumes
$\varepsilon^{F}_{t}$ at world time $t$ regardless of its own generation index.

\item[Definition 3 (Verifiable spatiotemporal locality \& outcome fidelity).]
Forking the simulator at $\tstar$ under $I$ yields the ground-truth
counterfactual render $y^{CF}$ and per-frame causal-descendant masks $D_t$.
Define
\begin{align*}
\text{locality violation } L &= \textstyle\sum_{t>\tstar}
  d\!\left( x^{CF}_t \odot (1-D_t),\; x^{F}_t \odot (1-D_t) \right),\\
\text{outcome fidelity } O &= \textstyle\sum_{t>\tstar}
  d\!\left( x^{CF}_t \odot D_t,\; y^{CF}_t \odot D_t \right),\\
\text{reward } R &= -(\lambda_O \cdot O + \lambda_L \cdot L).
\end{align*}
Both terms are computable without a learned judge.
\end{description}

\paragraph{Remark (exact abduction).}
Since $\tau^{F}$ is generated by the model, $\varepsilon^{F}$ is stored rather
than inferred; Pearl's abduction step is exact. In contrast, inversion-based
pipelines recover exogenous noise only approximately, and identifiability of
exogenous noise in high-dimensional generative models remains open~\cite{causaladapter2025}. Noise
coupling is what converts minimal-change from a distributional criterion into a
per-sample testable one.

\paragraph{Remark (operationalization).}
$D_t$ is computed from the paired ground-truth branches on the channels the
simulator exports: the per-pixel difference channel always, and an entity-level
channel where stable entity identities are available. On frames where the
complement of $D_t$ is empty --- in particular under interventions whose
viewpoint change covers the frame --- $L$ is \emph{undefined rather than zero},
and we therefore report $L$ stratified by intervention class. Pixel comparisons
exclude fixed overlay regions and apply a small magnitude threshold to absorb
characterized renderer noise, which is attributable to neither a locality
violation nor sampling stochasticity.

\section{A minimal illustration}

\begin{figure}[t]
\centering
\includegraphics[width=\linewidth]{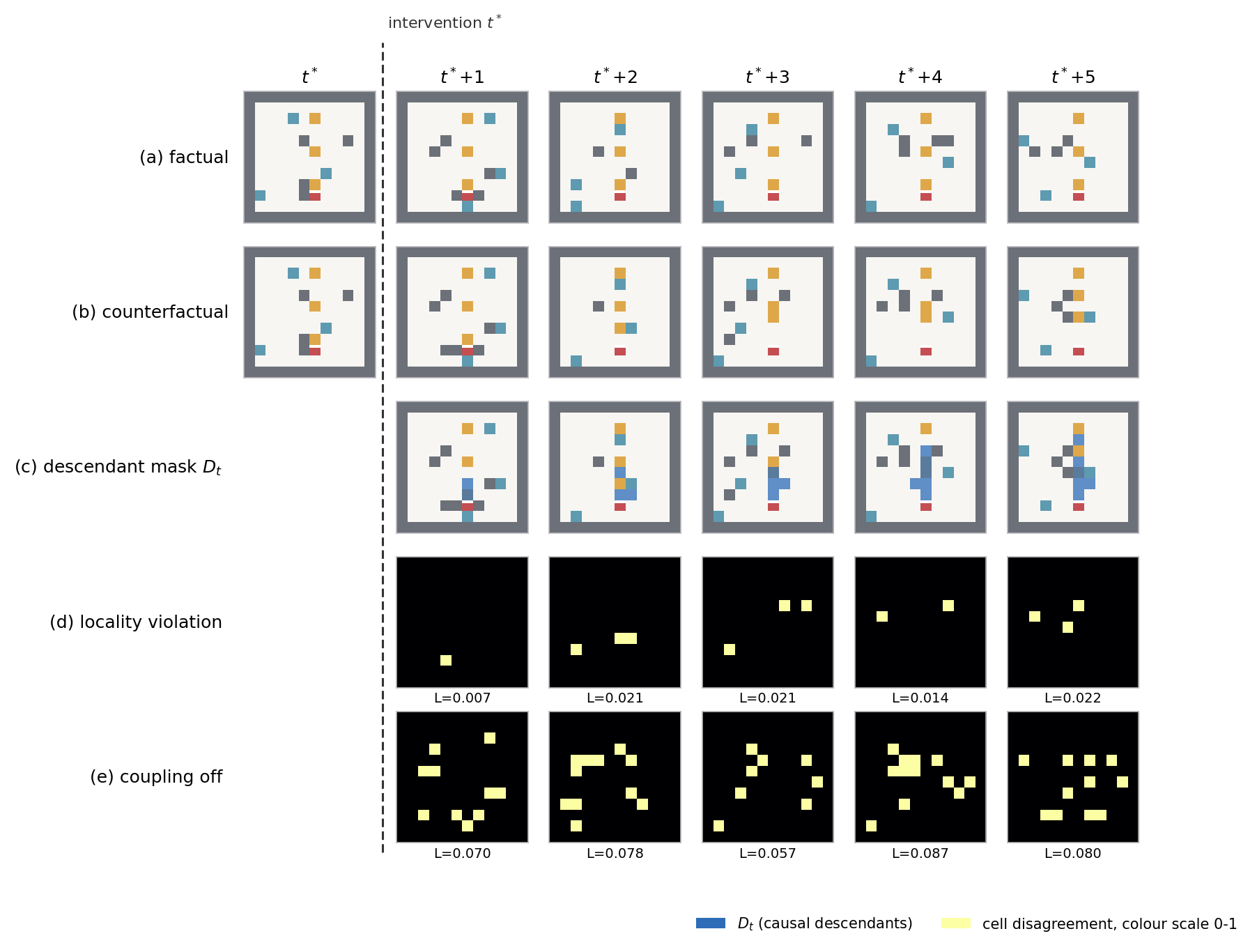}
\caption{Noise-coupled twin rollouts in an interactive world model, instantiated
in a deterministic grid world whose descendant masks are exact by construction.
Rows (a) and (b) share the self-generated prefix and the future exogenous noise
sequence, diverging only in the action stream from the intervention point
$\tstar$. Row (c) shows the causal-descendant mask $D_t$ obtained by forking the
simulator at $\tstar$. Row (d) shows the locality violation outside $D_t$; under
noise coupling, any signal there is attributable to a locality violation rather
than to sampling stochasticity. Row (e) repeats row (d) with the noise coupling
removed, where the same measurement is no longer possible per sample.}
\label{fig:twin}
\end{figure}

Figure~\ref{fig:twin} instantiates the construction in a small deterministic
environment as an existence proof: the twin branches are generated by a trained
autoregressive world model from a shared prefix and shared per-position sampling
variates; the descendant masks come from the simulator fork. With coupling on,
the region outside $D_t$ is nearly silent, and what remains is a genuine
locality violation of the model. With coupling off, the same region lights up
with sampling stochasticity, and the per-sample attribution of Definition~3 is
no longer available. Scale-up to a full interactive video world model, together
with the data engine and post-training results, is deferred to the next
revision.

\section{Positioning}

\paragraph{vs.\ CWMDT.}
CWMDT~\cite{cwmdt2025} formalizes counterfactual world models that accept interventions as
explicit inputs, conditioning a video diffusion model on LLM-edited
digital-twin text derived from an \emph{observed} scene. This is an open-loop,
single-shot setting: the factual video is externally given, interventions are
semantic attribute edits, and generation occurs once. We study the complementary
\emph{closed-loop autoregressive} setting: the factual trajectory is generated
by the model itself under an action stream, interventions are applied
\emph{mid-rollout} to actions or state, and the factual/counterfactual branches
are coupled through shared future exogenous noise --- making abduction exact
rather than approximated through an inversion or reasoning pipeline.

\paragraph{vs.\ What-If World.}
What-If World~\cite{whatifworld2026} contributes an evaluation benchmark for \emph{contrastive
intervention following}: paired test instances anchored on real frames,
differing in a single input-level variable, scored by the APEO protocol over
black-box generations. Our work differs on three axes: (i) we contribute a
generation method and a training objective, not an evaluation suite; (ii) our
interventions occur mid-rollout on the action stream rather than at the input;
(iii) our locality metric is defined under \emph{controlled exogenous noise}
--- a condition unenforceable when evaluating black-box models --- which
converts minimal-change from a distributional criterion into a per-sample
verifiable one.

\paragraph{vs.\ image-domain counterfactual diffusion and CSVC.}
A mature line of work instantiates Pearl's abduction--action--prediction
procedure in diffusion models via DDIM inversion for images~\cite{diffscm2022,semanticabduction2025,causaladapter2025}, and via
prompt-steered black-box editing for observed videos (CSVC~\cite{csvc2025}). We claim no
novelty for this correspondence. Our departure is the setting: no observed
artifact is inverted; the world model's own rollout supplies the factual branch
with exactly known noise, and the counterfactual is a \emph{branch of the same
generative process}, not an edit of a given video.

\paragraph{vs.\ PersistWorld.}
PersistWorld~\cite{persistworld2026} branches multiple candidate continuations from a frozen prefix to
stabilize RL post-training; its branches share the \emph{action} sequence and
differ in sampling stochasticity. Our construction is the exact dual: branches
share the \emph{noise} sequence and differ in actions. The former averages over
noise to stabilize training; the latter controls noise to expose the causal
effect of actions.

\paragraph{vs.\ the counterfactual-controllability framework.}
Concurrent work~\cite{cfctrl2026} articulates counterfactual controllability as a design
criterion for self-evolving world models at the framework level, without a
concrete generation or training procedure. We provide, to our knowledge, the
first method-level instantiation with a verifiable training objective.

\paragraph{vs.\ CounterScene.}
CounterScene~\cite{counterscene2026} performs counterfactual guidance for safety-critical scenario
generation in driving-specific BEV world models with predefined
agent-interaction graphs. Our formulation is domain-general over interactive
video world models and requires no hand-specified interaction structure;
descendant sets are obtained from simulator ground truth.

\paragraph{vs.\ the RLVR-for-world-models line.}
RL post-training of video world models with verifiable rewards is established~\cite{rlvrworld2025},
but existing rewards target reconstruction, perceptual similarity, geometric
consistency, or action following --- all properties of a \emph{single factual}
rollout. Ours is, to our knowledge, the first reward defined on
\emph{factual--counterfactual pairs}, directly optimizing counterfactual
correctness and locality.

\section{Limitations}

The locality metric inherits the capabilities of the simulator that grounds it.
Under egocentric cameras, interventions that move the viewpoint drive the
pixel-difference channel of $D_t$ toward full-frame coverage, rendering $L$
vacuous for that intervention class; we therefore stratify all locality
reporting by intervention class, treat empty-complement frames as undefined,
and confine training signal to viewpoint-preserving interventions until
camera-compensated comparison is validated. Renderer nondeterminism, where
present, is handled by fixed-region masking and magnitude thresholds rather
than assumed away. The illustration in Figure~\ref{fig:twin} is an existence
proof in a deliberately small environment, not evidence of scale.

\bibliographystyle{plain}
\bibliography{refs}

\end{document}